# Enabling scalable clinical interpretation of ML-based phenotypes using real world data

Owen Parsons[1], Nathan E Barlow[1], Janie Baxter[1], Karen Paraschin[2], Andrea Derix[2], Peter Hein*[2], Robert Dürichen*[1]

[1] Sensyne Health, Oxford, UK; [2] Research and Development, Pharmaceuticals, Bayer AG, Wuppertal, Germany

Contributions: (I) Conception and design: all authors; (II) Administrative support: AD, RD; (III) Provision of study materials or patients: RD; (IV) Collection and assembly of data: OP, NB; (V) Data analysis and interpretation: OP, NB, JB, KP, PH, RD; (VI) Manuscript writing: all authors; (VII) Final approval of manuscript: all authors

*Correspondence to: Robert Dürichen (robert.durichen@sensynehealth.com) and Peter Hein (peter.hein@bayer.com)

**Background**: The availability of large and deep electronic healthcare records (EHR) datasets has the potential to enable a better understanding of real-world patient journeys, and to identify subgroups of patients currently grouped with a common disease label but differing in outcomes and remaining medical need. ML-based aggregation of EHR data is mostly tool-driven, i.e., building on available or newly developed methods. However, these methods, their input requirements, and, importantly, resulting output are frequently difficult to interpret, especially without in-depth data science or statistical training. This endangers the final step of analysis where an actionable and clinically meaningful interpretation is needed.

**Methods**: We conducted a patient stratification and sub-phenotyping of cardiovascular patients from combination of NHS EHR datasets with encounters from March 2014 to August 2020 from Oxford University Hospital (OUH) and from February 2014 to March 2020 from Chelsea Hospital and Westminster Hospital (ChelWest). The dataset contained 1480 and 918 (aged 18 or older) from OUH and ChelWest respectively with EHR data containing diagnoses, laboratory tests, medications, and procedures. The use of different clustering methods and preprocessing steps resulted in more than 100 clinical reports which would require enormous amounts of time to be evaluated by clinical researchers.

**Results**: We have developed several tools to facilitate the clinical evaluation and interpretation of unsupervised patient stratification results, namely *pattern screening*, *meta clustering*, *surrogate modeling*, and *curation*. These tools can be used at different stages within the analysis. As compared to a standard analysis approach, we demonstrate the ability to condense results and optimize analysis time. In the case of *meta clustering*, we demonstrate that the number of patient clusters can be reduced from 72 to 3 in one example. In another stratification result, by using *surrogate models*, we could quickly identify that heart failure patients were stratified if "blood sodium" measurements were available. As this is a routine measurement performed for all patients with heart failure, this indicated a data bias. By using further *cohort* and *feature curation*, these patients and other irrelevant features could be removed to increase the clinical meaningfulness.

**Conclusions**: This study investigates approaches to perform patient stratification analysis at scale using large EHR datasets and multiple clustering methods for clinical research. We show examples on the effectiveness of the methods and hope to encourage further research in this field.

# Introduction

Research in the life sciences has come to rely heavily on large scale digital data acquisition and analysis.(1) Indeed, digitalization in health care, and specifically documentation of electronic health records (EHRs), is developing into a standard practice for many health care providers. The increased availability of large clinical datasets, combined with recent advances in machine learning methods, have led to an increasing number of studies.(2) Although this trend started over a decade ago, the available data is becoming more and more relevant as datasets are a) more complete, b) extended more longitudinally, and c) more horizontally integrated with increased linkage to other relevant datasets from the same patients like laboratory values, imaging, and other diagnostic procedures. These advances have underpinned the growing interest in the application of this large-scale real-world data not only for epidemiological purposes, but also to understand trajectories of patient subgroups within large but heterogenous diseases like heart failure, chronic kidney disease, or stroke. As the trend towards more complete EHR dataset is accompanied by development of analysis methods (e.g., machine learning), this approach is becoming more likely to reveal relevant and actionable insights that have the power to benefit patients directly (e.g., through better care and precision medicine) and indirectly (e.g., by facilitating development of tailored drugs).(3)(4)(5)(6)

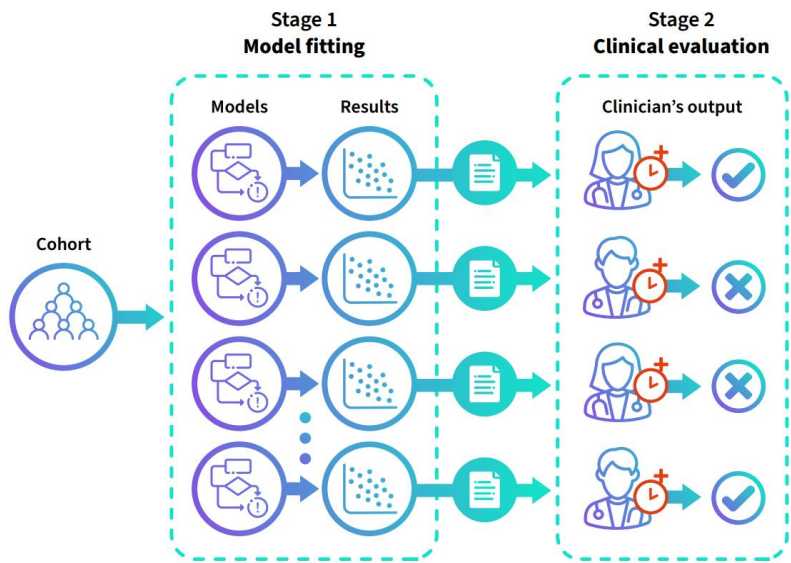

*Figure 1: General workflow of patient stratification where multiple models are fit based on a single patient cohort. The outcomes are validated by a domain expert in parallel for each model experiment, which is highly labour-intensive.*

One high impact area of research is patient stratification where, amongst others unsupervised learning methods, patients that share a similar clinical history (e.g., similar comorbidities) are clustered into sub-phenotypes to support disease understanding and facilitate more targeted treatment options.(7)(8)(9)(10)(11)(12)(13) This has been applied to problems such as identifying subgroups of intensive care patients with common clinical needs(14)(15) as well as finding subgroups of patients that have distinct responses to a fixed treatment.(16) Generally, there are two core stages to the process. The first is the identification of patient subgroups using data-driven methods. The second is the clinical evaluation and interpretation of these subgroups using statistical methods. New studies in this area often focus on the development and application of novel clustering methodologies,(17) however exploration of novel tools to facilitate and accelerate the clinical evaluation are not considered to the same extent. This development results in an increasing number of methodological choices and often it is impossible to determine initially which approach will lead to the most insightful outcome. Consequently, multiple approaches are applied which results in an increased number of potentially relevant outcomes to be evaluated by clinical experts (see Figure 1). As dataset sizes and number of model experiments increase, there is a growing need for novel tools that specifically support interpreting results from complex studies, where many parallel approaches to handling clinical evaluation output are applied.

In this publication we present several newly developed clinical evaluation tools that address the challenges posed when conducting large-scale, machine learning driven clinical analyses of unsupervised patient stratification. At a high-level, these challenges break down into managing large volumes of evaluation results that need to be interpreted by clinicians, facilitating the extraction of insights in studies with large number of observations and support fast iterations of results to increase clinical relevance. The solution we propose (Figure 2) extends the clinical evaluation process with the addition of a *key results identification stage* (stage 2), an *explainability stage* (stage 3), and an *optimisation loop* (stage 5). Note, even though not every tool presented is new, we focus in this publication on how the discovery process using ML methods and large scale EHR datasets can be improved by reducing the burden on clinical domain experts. We hope that this initiates further discussion within the community about the development of more appropriate tools.

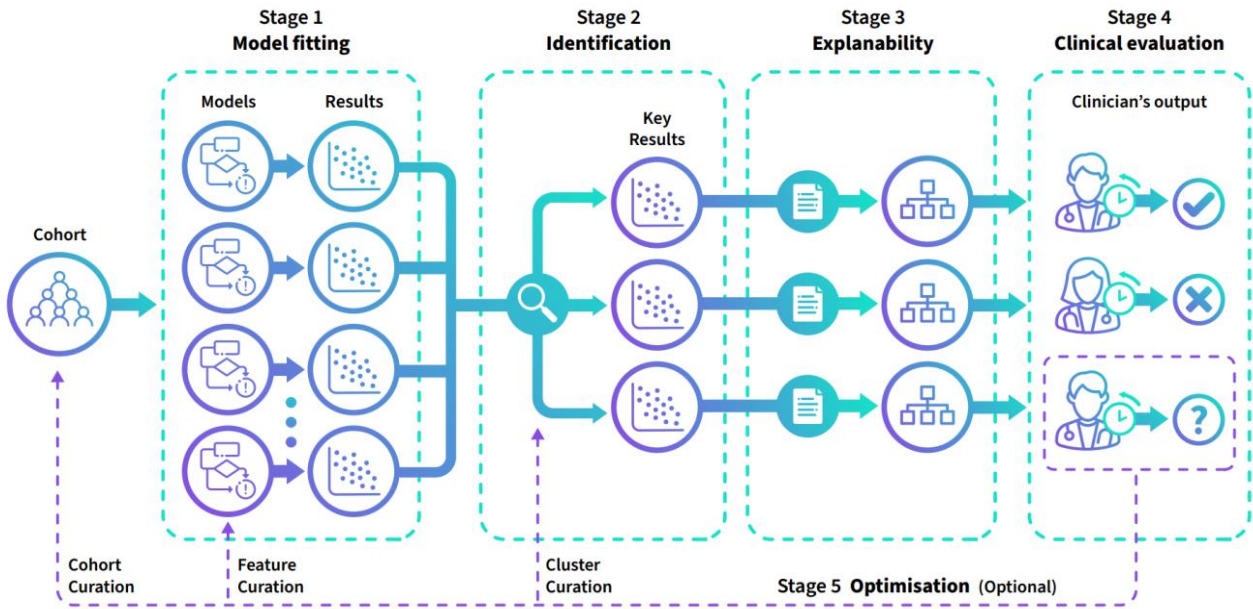

*Figure 2: Proposed clinical interpretation approach which allows a faster evaluation and iteration of ML-derived patient clusters to reduce the time burden on clinical researchers. By introducing the identification (stage 2), the explainability (stage 3), and optimisation stage (stage 5) the time required to evaluate a single set of results will be reduced dramatically in stage 4.*

# Methods

We present the conceptual approach and methodological details of our proposed new scalable clinical evaluation framework including methods for new stages. Additionally, we provide details of a use case study, along with the objectives of the study and the analytical approaches taken.

***Proposed scalable clinical interpretation approach***

In this publication we propose to extend the current two-step workflow Figure 1 by three additional stages with the objectives of a) reducing the number of results a clinical researcher needs to investigate, b) increasing explainability of individual results, and c) providing tools to quickly iterate and optimize analysis. Importantly, we suggest that this extension makes the clinical interpretation *scalable*. Our proposed workflow is shown in Figure 2. Note, the original two stages presented in Figure 1 remain the same (model fitting [stage 1] and clinical evaluation [now stage 4]). However, with the introduction of the *Identification, Explainability* and *Optimisation* stages the overall time required for a clinical researcher to review results will be reduced.

Stage 2 – Identification

The main objective of this stage is to reduce the number of available results generated in stage 1. As motivated in the introduction, for any patient stratification analysis, there exists an extensive number of potential experiments and often it is not possible to determine *a priori* which setting is most appropriate for analysis. Some of the analytical choices to be considered are:

- *Data types*: As modern EHR datasets become more complete, multiple data types such as diagnosis codes, laboratory values, oncology medication, or clinical reports become available. Even though ideally as much information as possible should be considered for the analysis, each data type might contain biases which could impact the stratification result.
- *Data preprocessing*: Information in EHR data can be preprocessed in different forms, some of the choices to be made are: handling of continuous data types (e.g. laboratory values could be coded in a binary format, only indicating presence or absence of a laboratory test, or using the raw values), filtering of data elements (e.g. removing of data elements which are less then *X*% present in the dataset to remove noise), or hierarchy adjustment of data elements (e.g. ICD10 based diagnosis codes have a multiple hierarchy levels)
- *Handling of temporal progression:* EHR data ranges across multiple years, consequently different types of analysis are possible e.g., considering only data recorded during a specific hospital admission or data across a specific time range.
- *Cluster methods:* As mentioned in the introduction there is a large number of cluster methods available ranging from simple k-means clustering(18) to more complex deep learning based clustering,(19) each with its advantages and disadvantages.
- *Expected numbers of clusters:* For many clustering methods, the number of expected clusters *k* in a dataset needs to be defined. As this is normally not known *a priori*, often multiple numbers of clusters are evaluated. Alternatively some cluster methods have built-in mechanism to find the *optimal* number of clusters based on a predefined criteria.(20)

A reduction in the number of experiments to be analyzed can be achieved through various approaches, such as: a) automatic screening of all results and identification of commonalities, e.g., certain patients are always grouped together, or b) automatic ranking with respect to the clinical objectives of the study and identification of the most relevant results. Two examples of such approaches are *meta clustering* and *pattern screening,* respectively.

**Meta clustering:** To easily extract the common trends across multiple experiments, we can turn to approaches that consider the consensus across several different sets of results. Indeed, from a clinical perspective, it is particularly useful to have access to both qualitative and quantitative measures of the extent of agreement between different clustering algorithms. We adopted a procedure for combining results from multiple analyses using meta consensus clustering (21). A previous example of the application of meta clustering for patient sub-phenotyping can be found from Aure *et al.*(22). In addition, we generated hierarchical clustering heatmaps, or dendrograms, which display the cluster assignments of all individual patients across the different experiments included in the meta consensus clustering analysis. These dendrograms can illustrate the degree of overlap in cluster assignments between distinct experiments.

Assuming *n* cluster analysis experiments were performed in stage 1 on a dataset with *p* patients, each individual patient $x_i$, with $i = \{1,...,p\}$, is assigned to a cluster $c_{i,j}$ for experiment *j* with $j=\{1,...,n\}$. The data is converted to a binary dataframe where the individual patients are represented along the rows and clusters for each individual experiment were represented across the columns (Figure 3a). This data is then clustered by agglomerative hierarchical clustering (23) using the average linkage technique which works by iteratively pairing and merging clusters until all clusters have been merged into a set of meta clusters $c^*_k$. We used the Hamming distance to quantify the degree of dissimilarity between patients in this cluster space. This generates a dendrogram representing a hierarchical structure of patient clusters that can be split at different levels of hierarchy to obtain different numbers of clusters (Figure 3b).

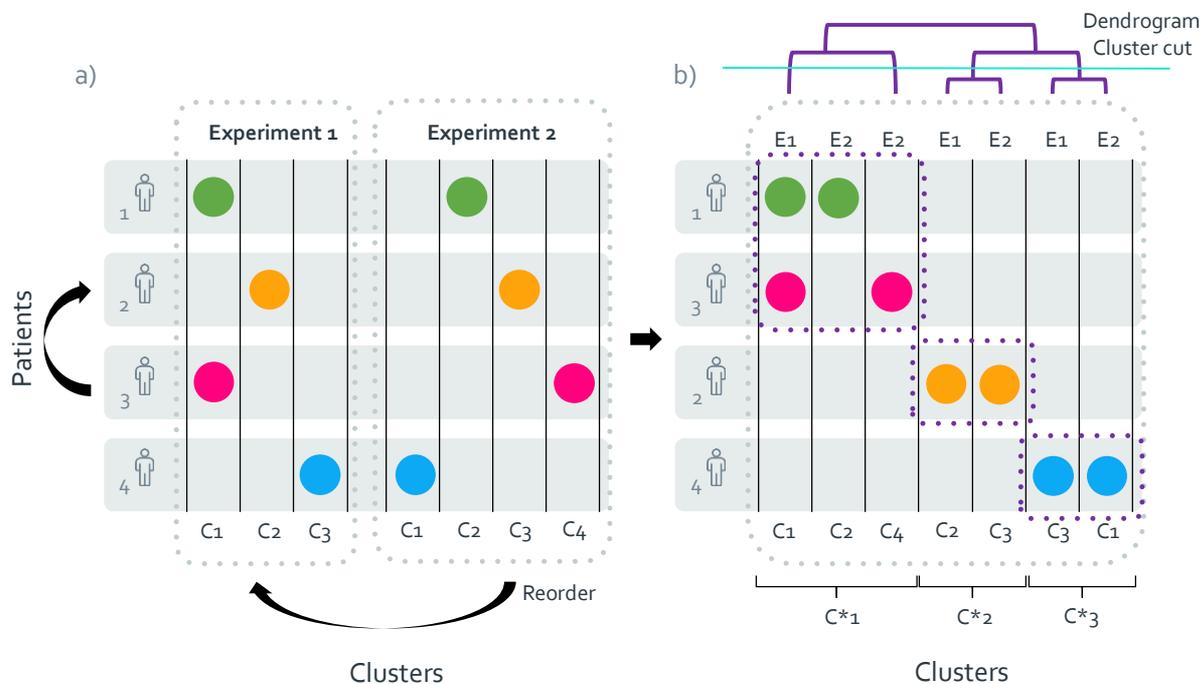

*Figure 3: Illustration of meta clustering approach, where two experiments E1 and E2 are combined by reordering the patients and cluster labels so that the similar patients are in proximity. The original set of experiments and clusters a) are reordered so that the new meta clusters b), denoted ban an asterisk * and bounded by the purple dotted lines, are now found to be C\*1: E1-C1, E2-C2, E2-C4; C\*2: E1-C2, E2-C3; C\*3: E1-C3, E2-C1. A dendrogram in purple above b) demonstrates how the clusters are split at each level of hierarchy. The height of the blue dendrogram cluster cut line controls the number of clusters. Note that the patients are colour coded for clarity.*

Note, also for the meta clustering approach, the number of clusters *k* needs to be defined. To address this, meta clustering was performed for $k = \{n : n \text{ is an integer with } 1 < n < 13\}$ clusters. Each set of meta clustering results was evaluated using the average silhouette index. Higher silhouette index values indicate better cluster separability, i.e., clustering quality.(24) Consequently, the numbers of clusters resulting in the first and second local maxima of silhouette index values are automatically selected to produce the results.

***Pattern Screening:*** Usually clinicians follow an intuitive, experience-based approach when evaluating clinical cohorts and their properties; naturally this limits throughput. Therefore, we developed a pattern screening tool for emulating a clinical review of experiment results. When evaluating results from multiple cluster analysis experiments, typically a reviewer will screen reports by applying a set of – often implicit – concepts derived to determine which are of clinical interest. The pattern screening process takes these concepts and implements simple algorithms, such as identifying clusters with an increased mortality risk, to automatically rank the set of results. Depending on the clinical objective, a wide range of rules can be defined to reflect the different study objectives. Note it is critical to determine rules before analysis to prevent the introduction of selection bias.

As outlined in the **Use-case method** section, the objective of our example study is to identify patient clusters which differ with respect to outcomes such as mortality. Additionally, the identified patient clusters should be easily explainable (see Section Stage 3 – Explainability). These objectives were translated into the following rules:

1. For all clusters *m* across the individual set of results *n*, compute the hazard ratio $HR_{m,n}$ using the "cluster vs rest" Cox proportional hazard model with respect to *mortality, recurrent stroke, bleeding events* and *re-hospitalisation*
2. Calculate the log rank p-value, $p_v$, from the Cox model
3. Repeat step 1 & 2 with patient clusters identified via the surrogate model (see section Stage 3 – Explainability)
4. Compute cluster and results specific ranking score $R_{m,n}$ by $R_{m,n} = -\log(p_v)$ and sort results
5. Compute the average between base model and surrogate model scores, $R_{m,n}^{average} = \frac{R_{m,n}^{surrogate} + R_{m_n}}{2}$

Stage 3 – Explainability

With hundreds of different data points captured in modern EHR systems, it becomes increasingly challenging to capture the key characteristics for a specific patient cluster (see Figure 4). The purpose of this stage is to translate complex patient clusters, which might be generated by black-box deep learning models, into understandable (and ideally explainable) results using surrogate models, such as decision trees, which provide explainability by design (25). This is a very active research field within the data science and ML community and a full presentation of all possible approaches is beyond the scope of the paper. Therefore, the explainability method employed in our analysis is chiefly a decision tree that is trained on the input model features and the cluster labels of the black-box model, which is referred to here going forward as the surrogate model.

**Surrogate models:** In general, there is a trade-off between model complexity and interpretability (26). Due to the increasing number of data points, there is a tendency to use more complex clustering models which makes rapid model interpretation difficult. Additionally, further tools might be required if consensus clustering approaches, e.g., meta clustering,(27) are used. This challenge is addressed by using surrogate models, which are secondary white-box models trained to predict the outputs of the more complex model.(28)(29)(30) Surrogate models give a more complete picture than enrichment analysis alone (See Figure 4). While enrichment is useful for finding features that are significantly more prevalent in each cluster, surrogate models can be used to understand feature interactions and specific feature thresholds that determine patient cluster assignments.

Using a surrogate model, the trained parameters can be used to interpret the extent to which individual features influenced the clustering process. In the case of patient stratification from EHR data, this means we can explain which medical features were most important in determining which cluster a patient should be assigned to. Through there are several methods available, we use primarily supervised decision trees in our surrogate model approach. Also, the ground truth prediction labels are defined as the one-hot-encoded cluster labels from our black-box clustering algorithm. For each of these methods, we trained a model using 5-fold cross-validation. This approach involves randomly splitting the dataset into 5 groups of equal size and then iteratively selecting each of these 5 groups to be the validation set while the remaining 4 groups were used as the training set. Unlike most machine learning analyses, we did not use an additional test set to evaluate the final model performance as the models were intended for purely explanatory purposes.

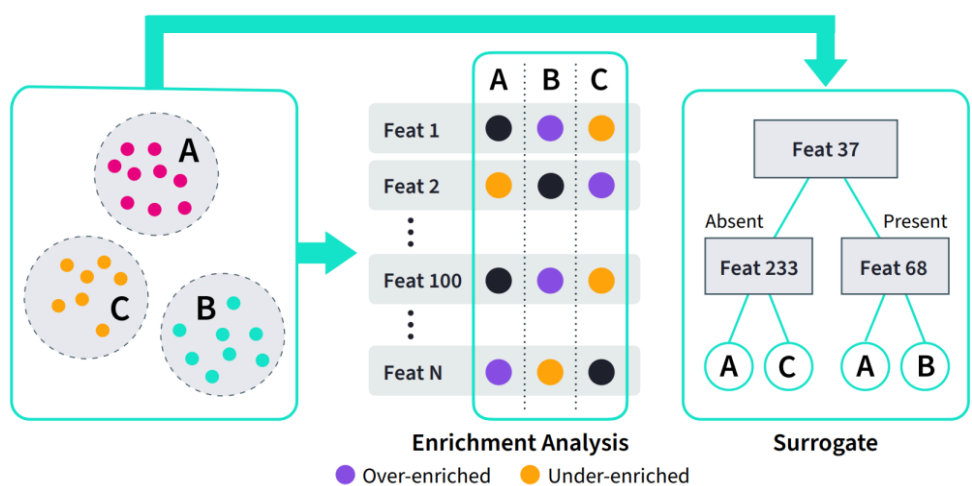

*Figure 4: An illustration of two methods for interpreting patient stratification results. A simple and direct approach is enrichment, where feature counts (and percent) for each cluster of patients is calculated. The odds ratio (OR) between enrichment across clusters indicates the extent of over-enrichment (OR>1) or under-enrichment (OR<1) of a feature. With an increasing number of features in modern EHR datasets, direct enrichment analysis is impractical and alternative approaches, e.g., surrogate modelling, should be applied. A supervised decision tree finds criteria for cluster labels based on the important features in the model. Consequently, only the most important features are used which vastly simplifies the analysis.*

Stage 5 – Optimisation
The purpose of this stage is to provide methods for iterative optimisation of patient stratification results. For example, a model may find a patient cluster with a very high risk of mortality – in line with the objective of the study – however, a surrogate model might indicate that the cluster is defined by non-clinically meaningful features. Therefore, tools to remove or modify features and/or patients to validate potential insights and safeguard against data bias are crucial.

We have developed methods which can be applied at different stages of the stratification analysis, namely to cohort, feature, or cluster curation, to enhance the quality of the cluster analysis (see Figure 2). Note that this stage is highly dependent on domain expert knowledge.

The process of curating either features or patient cohorts leads to changes in the model input data which necessitates the clustering process to be re-run prior to clinical evaluation. Curating the clusters directly – i.e., before review and analysis – means the input data does not change and the clinical evaluation can be performed directly.

**Cohort Curation** – the objective of cohort curation is to remove or add additional patients from the cohort of interest. This might sometimes be required as an initial cluster analysis could reveal patient clusters which are defined by a previously unknown data bias, e.g., a patient cluster contains only patients before a specific year due to a change in standard practices over time, such as the brain natriuretic peptide (BNP) test versus the later introduced NT-proBNP test for heart failure. Naturally, such patient groupings typically have poor clinically utility in terms of providing meaningful stratification despite formally meeting the criteria for inclusion in the cohort. In this case it can be beneficial to further exclude some patients before clustering analysis is re-run, thereby avoiding the risk of diluting potentially relevant signals in the data.

**Feature Curation** – As real world EHR data tends to be messy and biased, there will be instances where clusters are defined based on clinically irrelevant or obvious features. The objective of feature curation is to either remove non-informative individual features or combine multiple non-clinically relevant features to a single clinical meaningful feature (see examples in Figure 5). This process strongly depends on domain expertise as deciding which feature is useful and not strongly depends on the objective of the analysis. Allowing the clinical researchers to quickly curate features and rerun the analysis will result in a clearer and more clinically relevant cluster definition. Note, that when feature curation is applied, the patient cohorts are unchanged, but the clustering analysis and subsequent clinical analysis are re-evaluated using the curated feature list.

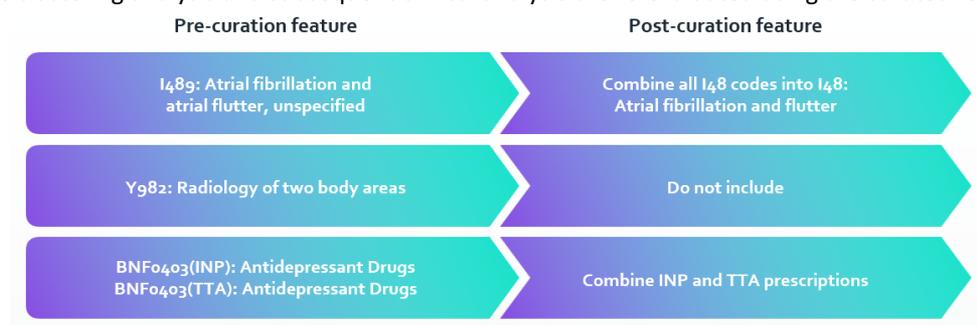

*Figure 5: Feature curation examples where diagnoses codes I48 are generalized, procedures codes Y982 are excluded and BNF medication codes are combined where the prescription type, inpatient (INP) versus to take away (TTA), is not clinically relevant.*

**Cluster Curation** – There are several reasons for identified clusters to be removed or to be combined or with other clusters. For example, if a cluster appears to suffer from bias (such as by data source), it can be removed or if two clusters appear to exhibit similar attributes then they can be combined. It is sometimes the case that clinical evaluation of clusters leads to two or more clusters being identified as having very similar survival or enrichment profiles. When this occurs, it can be useful to combine these clusters into a single cluster. Similarly, if there are multiple clusters that are not informative or relevant to a given clinical question then it can be useful to condense these less relevant clusters into a single cluster.

As cluster curation involves manually adjusting the cluster definitions after the cluster analysis, it is not required to re-run the clustering for this type of curation to be applied. Instead, just the clinical evaluation stage is re-run with the expectation that the results from these analyses are easier to interpret due to the reduced number of clusters and therefore a smaller number of comparisons.

***Use-case methods (Dataset, clinical questions, and analytical approach)***

To illustrate the challenges with large scale patient stratification studies and the advantages of our proposed approaches, we use results of a stratification study within the cardiovascular domain with a specific focus on stroke and heart failure patients. In the following section, the EHR dataset, clinical questions, cohorts, and analytical approaches will be described.

Dataset

The dataset used for our analysis consisted of anonymized EHRs obtained from two NHS trusts, Oxford University Hospital (OUH) and Chelsea and Westminster Hospital (ChelWest). The total number of patients available in the study was 860,545, with records spanning from August 2010 to March 2020.

The dataset contains 6 different types of clinical features: diagnosis codes (ICD-10 codes), procedure codes (OPCS-4 codes), medication codes (BNF codes), laboratory values, demographic information, and administrative information. Laboratory values are mainly continuous, while diagnosis, procedure, and medication codes are binary or categorical features (indicating presence or absence). Administrative information contained data such as start and end date of admissions and admission type (e.g., inpatient or outpatient). Diagnosis codes could appear either as a primary diagnosis (in the case that the diagnosis was the primary reason for the hospital admission) or secondary diagnosis (in the case that the diagnosis was a comorbidity).

The methodology developed and presented here can be used to address explorative questions, such as: can unbiased sub-phenotypes be identified for a given condition? The examples given below relate to identifying sub-phenotypes of patients who have a first diagnosis of ischemic stroke or heart failure within the available data set.

Cohort Definitions

Based on the clinical question above, relevant criteria were applied to the starting cohort 608,759 (OUH) and 251,786 (ChelWest), which left 1430 (OUH) and 1062 (ChelWest) heart failure patients, and 1480 (OUH) and 916 (ChelWest) stroke patients (Table 1). To obtain these sub-populations of patients with relevant medical profiles for our analysis, we defined the cohorts by an index date for each patient which is the first diagnosed acute event in their disease course. For the heart failure cohort, patients were included in the cohort if they had a heart failure event, which was defined as the occurrence of any of the following ICD-10 codes as a primary diagnosis:

- I50* Heart failure
- I11.0 Hypertensive heart disease with (congestive) heart failure
- I13.0 Hypertensive heart and renal disease with (congestive) heart failure
- I13.2 Hypertensive heart and renal disease with both (congestive) heart failure and renal failure.

Patients were only included in the cohort if they had one of these events as well as at least 3 months of medical data available prior to the first admission where one of these diagnosis codes was recorded. We also excluded patients from the cohort based on several additional criteria, defined as:

- Patients whose first admission (as defined above) was under 48 hours and had a heart failure related procedure code in the 30 days following the first admission (OPCS-4 codes: K59*, K60*, K61*, K72*, K73*, K74*)
- Patients who had heart failure related ICD-10 codes recorded as a secondary diagnosis prior to their first admission (ICD-10 codes: I50*, I11.0, I13.0, I13.2)
- Patients who had been prescribed Eplerenone, Sacubitril with Valsartan or Spironolactone at a dose of either 25mg or 50mg prior to their first admission
- Patients with a recorded New York Heart Association classification or patients with a recorded ejection fraction under 40% prior to their first admission

For the stroke cohort, we included patients who met one of the two following criteria:

- Patients aged 18 or older who had an admission for ischaemic stroke as a primary ICD-10 code (I63* Cerebral infarction) and had at least 6 months of medical data available prior to the first admission where they had this event
- Patients with a primary or secondary I63* Cerebral infarction ICD-10 code OR I69.3 Sequelae of cerebral infarction ICD-10 code that occurred prior to the first record of an ischaemic stroke admission (as defined above)

The outcomes considered for these patients were mortality and for stroke patients recurrence of stroke and bleeding. Readmission for heart failure was also considered. The definitions for these end points are available in Appendix A.

The above cohort criteria quite significantly decrease the number of patients available for analysis, all cohorts are less than 0.5% of their original (Table 1). With respect to cohort features, a 1% filter was applied that ensured all patients had feature coverage greater than 1%, e.g., at least 1% of possible features were present for a given patient. The criteria (unsurprisingly) increase the percent of patients with an outcome of mortality, where the original cohorts had 6-8% mortality, the sub-cohorts ranged from 30-47%. Note that the heart failure and stroke cohort demographics maintained an even sex balance, however the average patient age increased from 50s to late 70s. In these instances, the patient age is defined as the difference between birth date and date of event. The mean patient clinical observations increased in the sub-cohort populations as many patients with a low number of observations were filtered out. The time coverage is highly dispersed and skewed – many patients have single day coverage; hence the standard deviation is larger than the mean. However, we are focusing on "at-event" and data outside the event will be disregarded. The discrepancy between the number of unique laboratory values in OUH vs ChelWest is simply a matter of tests available at given trust. Furthermore, there is only a subset of LIMS that a relevant for the cardiovascular patient stratification study.

*Table 1: Table of patient data breakdown from Oxford University Hospital and Chelsea And Westminster Hospital for ischaemic stroke, and acute heart failure.*

|  | Source cardiovascular dataset | | At Heart Failure Event | | At Stroke Event | |
| --- | --- | --- | --- | --- | --- | --- |
|  | **OUH** | **ChelWest** | **OUH** | **ChelWest** | **OUH** | **ChelWest** |
| Total Patients (860,545) | 608,759 | 251,786 | 1430 | 1062 | 1480 | 916 |
| Outcome Mortality | 38,039 | 21,028 | 516 | 505 | 448 | 389 |
| Demographic<br>• Age (years) - mean (std)<br>• Female<br>• Male | 53.1 (20.0)<br>328,760F<br>279,885M | 59.9 (20.2)<br>133,535F<br>118,075M | 78.1 (13.5)<br>688F<br>742M | 78.7 (11.6)<br>533F<br>529M | 77.6 (12.8)<br>769F<br>711M | 77.1 (12.9)<br>469F<br>447M |
| Clinical observations – mean (std) | 39.1 (27.5) | 65.1 (40.7) | 75.4 (33.7) | 90.0 (38.9) | 77.2 (19.0) | 93.2 (37.5) |
| Time coverage (days) – mean (std) | 675 (637) | 762 (1,097) | 506 (453) | 605 (680) | 462 (442) | 560 (711) |
| No. of unique diagnosis codes | 10,800 | 8,907 | 1,064 | 1,137 | 1,186 | 1,219 |
| No. of unique procedure codes | 6,793 | 4,758 | 173 | 229 | 225 | 211 |
| No. of unique medication (BNF) codes | 14,372 | 3,619 | 116 | 105 | 116 | 116 |
| No. of unique laboratory codes (LIMS) or (vitals) | 90 | 1,599 | 47 | 123 | 45 | 123 |

Data Pre-processing

The data preprocessing step consists of cleaning, quality checks, standardizing, time interval aggregating, and converting to a more useful data representation.

**Cleaning and Quality Checks**: There are several sources of errors and irregularities in raw datasets, such as missing values, mismatched data, typos, superfluous formatting, and duplication. Quality checks are designed to catch these types of problems, in particular demographic data is checked that the age of the patient greater than 18 and that the age of death is less than the age at first admission. Further, demographic and laboratory measurements are checked to be within physiological ranges and that the values are of correct type and not empty.

**Standardization**: The cleaned data is then standardized across trusts. The standardization includes variations in naming convection, e.g., 'Amikacin', 'AMIKACIN', 'AMIK', 'ARM', 'AMIKCAIN LEVEL', as well as by units, e.g., 'g/l' vs 'mg/dl' which must be scaled. Additionally, granular subcategories are standardized, such as the mapping of medications to a parent code, e.g., 'Timolol', 'Pilocarpine Nitrate', 'Pilocarpine Hydrochloride' to 'Treatment of Glaucoma'. The structure of the raw data was also harmonized, for example, data columns that were spread across multiple columns were combined to a single column.

**Aggregation**: As the raw EHR data consists of datapoints across time, it is necessary to aggregate and simplify the time intervals. For continuous variables, the observations within a time window are summarized using medians, median absolute deviation (MAD), count, minimum, maximum, and last observed value. For categorical variables, the observations are condensed to unique values or counts for each feature. Data was filtered by encounters for patients that contain the event of interest where the start and end dates are directly extracted from the EHR.

**Filtering:** Patients were filtered for sparsity. If a patient had a feature density of less than 1%, they were excluded from the cohort. Specifically, this was applied for patients with all possible laboratory test, medication, procedure, and diagnoses.

**Data Representation**: Before clustering, the preprocessed data is transformed into an *embedding* space which is more amenable to modelling. There are three types of data transformations used in this method, one-hot-encoding (OHE), 'GloVe' embedding, and quantisation. The simplest embedding method is OHE where the presence or absence of a feature is denoted in a binary fashion as a 1 or 0. The more complex 'GloVe' embedding(31) is a common natural language processing (NLP) technique which learns a dense vector representation for words trained on the word-word co-occurrences in a document (or *corpus*). Quantisation is a method used for handling continuous values, such as found in laboratory tests. In short, quantisation puts the continuous values into bins, quantising the values within a bin to a single value. In particular, we used distribution aware quantisation(32), where the number of data points in each quantisation bin are consistent across all bins. Note that we can also combine quantisation with GloVe.

Clustering Methods

Prior to the clinical evaluation, clustering analysis was performed on the heart failure and stroke cohorts with three different embeddings. We ran experiments for two clustering methods described briefly below (6 experiments in total; 3 different embeddings and 2 clustering methods). The first method is Deep Embedded Clustering (DEC). This technique was based on the work of Junyuan *et al* and consists of an auto-encoder with an additional clustering layer.(33) Similarly, a Modified Variational Autoencoder (MVA) was deployed. This method uses a modified KL-divergence term in the loss function during training which encourages separation of the individual clusters within the embedding space.(19) The optimum number of clusters k was determined using a bootstrapping approach. Briefly, the bootstrapping approach consists of reference models and subset models. The reference models are the trained unsupervised clustering models with all data for each cluster $k = \{3, \ldots, 10\}$. The subset models are generated from 10 subsets where each division represents 75% of the data sampled at random. The optimum cluster k is defined as the cluster number with the highest agreements between subset and reference models. The agreement is calculated as the Jaccard index,

$$J(C_1, C_2) = \frac{1}{N} \sum_{i=1}^{k} \sum_{j=1}^{l} n_{ij} \frac{n_{ij}}{m_{ij}}$$

Where $N$ is the number of datapoints in both clusters, $n_{ij}$ is the number of datapoints in cluster 1 and cluster 2, $m_{ij}$ is the number of datapoints in cluster 1 or cluster 2, and k and l are the number of clusters in the cluster results for $C_1$ and $C_2$.

Classical clinical evaluation

*Enrichment Methods*

Enrichment analysis is a useful tool for inferring feature importance from clustering results.(34) In particular, enrichment tables can be generated which summarize the extent to which patients adhere to a subgroup relative to the rest of the cohort. For categorical features, we measure the total and frequency (percent) in each group. For continuous features, we measure the mean and standard deviation as well as median and interquartile range for each group. We further employ statistical analysis by calculating a p-value (Fisher's exact test or Chi-squared if all four counts if the contingency table had values greater than 10), and the odds ratio. Note that the odds ratio is simply calculated from a contingency table and is informative of the direction of the enrichment, i.e., odds ratio values greater than 1 are *over-enriched*, and less than 1 are *under-enriched*. In the case of continuous feature enrichment, such as laboratory values, numerical features were evaluated using the Kruskal-Wallis test. Note that we defined a p-value threshold (adjusted using the Benjamini Hochberg method) of < 0.05 to determine significance.

*Kaplan Meier and Cox*

The survival analysis for each cluster was performed on the right-censored time to event data. The Kaplan Meier (KaplanMeierFitter) curves and Cox (CoxPHFitter) proportional hazard values were calculated along with the 95% confidence intervals using Lifelines.(35) The Cox models were adjusted for age and sex, the baseline hazard was calculated using Breslow's method, and ties were handled using Efron's method. No regularization was applied

*Software*

The models were developed in Python (RRID:SCR_008394) using the standard open-source packages: Pandas (RRID:SCR_018214), Numpy (RRID:SCR_008633), Scikit-Learn (RRID:SCR_002577), Matplotlib (RRID:SCR_008624), and Tensorflow (RRID:SCR_016345). Additionally the statistics and survival analysis was performed using Lifelines.(35)

# Results

In the following four sections, results will be presented for the meta clustering, pattern screening, surrogate modelling, and curation process. These results are generated using the patient stratification task outlined above. To demonstrate the power of this process, we illustrate with examples on a clinical study using a large-scale EHR dataset which focuses on the identification of novel heart failure and stroke sub-phenotypes. Note, the focus of this section is not to examine clinical details of the results which have been identified, rather to investigate how the clinical evaluation process was impacted by the proposed methods. To avoid confusion, the individual experiments generated for the different scenarios were labelled numerically, and experiments reviewed in this section using meta clustering are labelled using letters A-K (Table 2).

Table 2: Experiments used for demonstration purposes in for meta clustering, pattern screening, surrogate modelling, and curation.

| Trust | Disease | Cluster | Experiment Name | Used in |
|---|---|---|---|---|
| OUH | HF | 3 | A | Meta Clustering |
| OUH | HF | 5 | B | Meta Clustering |
| OUH | HF | 2 | C | Meta Clustering + Pattern Screening |
| OUH | Stroke | 3 | D | Pattern Screening |
| ChelWest | Stroke | 3 | E | Pattern Screening |
| ChelWest | Stroke | 6 | F | Pattern Screening |
| ChelWest | HF | 3 | G | Surrogate |
| OUH | HF | 6 | H | Feature & Cohort Curation (original) |
| OUH | HF | 3 | I | Feature & Cohort Curation |
| OUH | Stroke | 5 | J | Feature Curation (original) |
| OUH | Stroke | 5 | K | Feature Curation |

*Meta clustering Heart Failure Experiments*

The benefit of meta clustering is illustrated by the heart failure use-case with data obtained from OUH as described in the methods section. As outlined in the data pre-processing section and the clustering methods section, 2 clustering algorithms and 3 different possible pre-processing steps were used. For each algorithm and pre-processing step, the cluster number *k* was iterated between 2 to 11. In theory, this would imply the generation of up to 60 reports (2 cluster methods X 3 pre-processing steps X 10 cluster combinations). By using the bootstrapping approach defined in the Clustering Methods section, the number of clusters *k* was reduced to 1 or 2 per experiment resulting in total in 10 different set of patient stratification results with 72 clusters (Table 3). Applying the meta clustering approach on this set of initial results leads to 2 meta clustering reports automatically determined by the silhouette score with *k=3* and *6* (see Figure 6). The different cluster algorithms and pre-processing steps resulted in similar clustering results which could be identified by our consensus approach as indicated by the dense purple patient clusters in the dendrograms.

Table 3: Overview of initially generated patient stratification results using different cluster algorithms, pre-processing steps, and numbers of clusters k before applying meta clustering.

| Algorithm | Type of pre-processing | Number of clusters k | Number of experiments (Total = 10) |
|---|---|---|---|
| DEC | OHE | 5 | 1 |
| VAE | OHE | 5, 8 | 2 |
| DEC | GloVe | 6, 9 | 2 |
| VAE | GloVe | 8 | 1 |
| DEC | Glove with quantisation | 7, 10 | 2 |
| VAE | Glove with quantisation | 5,9 | 2 |

As an aside, another useful aspect of this approach is the ability to tune the number of clusters, whereby increasing the number of clusters, cluster groupings are further separated in a hierarchical manner. This is visible for Experiment A cluster 3 (*k=3*) (Figure 6 left), which in Experiment B is split into cluster 3 and 5 (*k=6*) (Figure 6 right). Note in case of Figure 6 right, a cluster of only 2 patients was identified which was excluded from the analysis to remove risk of patient identification. These patients are noted as "unclustered".

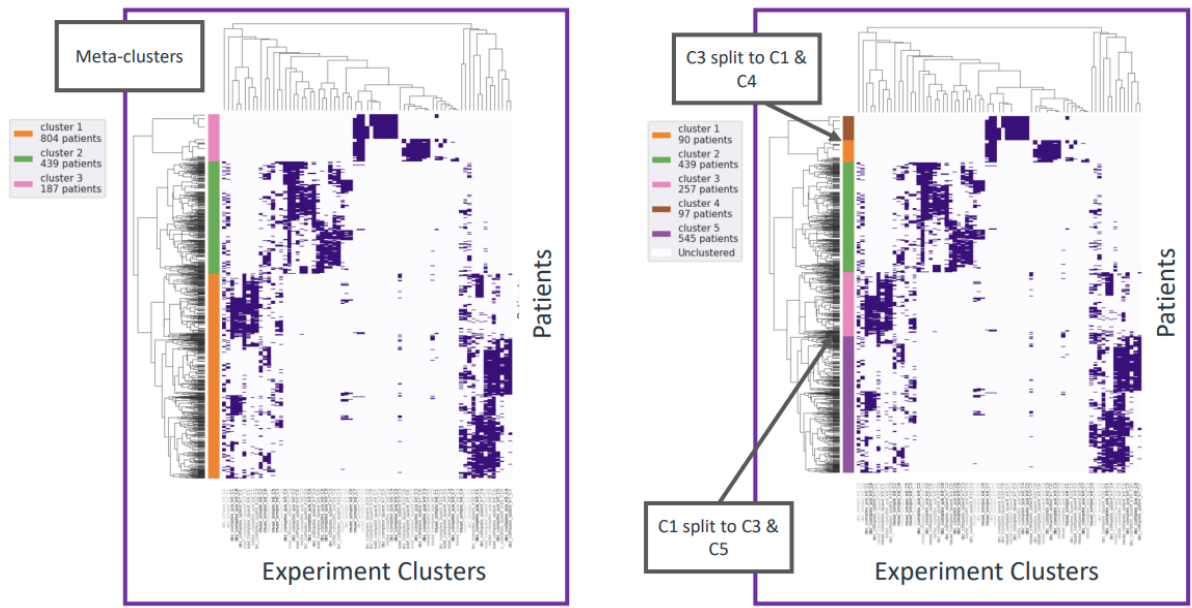

*Figure 6: Meta clustering results using the heart failure use case and data from OUH – Experiment A on the left and Experiment B on the right. The method identified similar cluster assignments reducing the number from 10 experiments (and 72 clusters) to 2 experiments with either 3 or 6 clusters.*

### Pattern Screening

The value of the pattern screening tool is illustrated on the stroke stratification example using results from both trusts OUH and ChelWest. Like the heart failure scenario, multiple encoding and cluster techniques were applied on 14 experiments: 7 cluster results for OUH and 7 for ChelWest. Additionally, meta clustering was applied to the results of OUH and ChelWest separately (experiment C-F). Using between 3 and 7 different numbers of clusters $k$, there were finally 71 sets of results to analyse. For each identified cluster, generated either by one of the cluster algorithms or meta clustering, a pattern screening score $R_{n,m}$ was computed for the outcomes: mortality, bleeding, and recurrent stroke. In the following example, we focus on mortality as the main outcome of interest, with bleeding and recurrent stroke scores used to supplement the analysis. To visualize these scores, we provide a heatmap of the mortality score for each experiment, which are placed in a scatterplot with the bleeding and recurrent stroke scores on the y- and x-axis, respectively. The experiment indices are arbitrary and are not significant to the example.

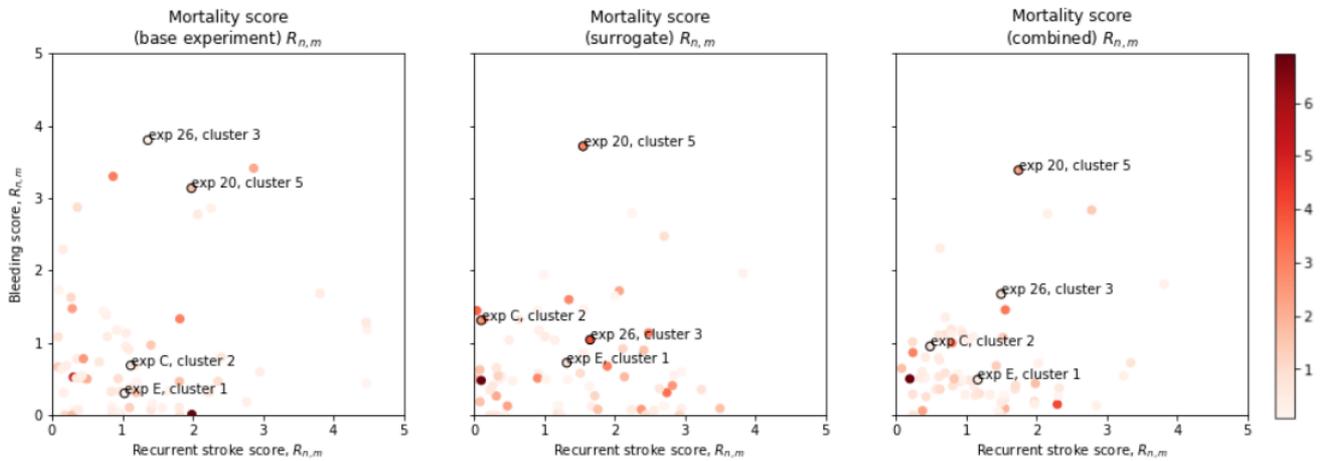

*Figure 7: Scatter plot heatmap of cluster specific pattern screening scores of the stroke stratification experiment using OUH and ChelWest data for the primary outcome mortality (red heatmap), and the secondary outcomes bleeding (y-axis) and recurrent stroke (x-axis). Scores range from 0 to infinity which correlates with an increasing significant difference between the cluster and the remaining patients of the cohort. The scores are calculated for the base experiment (left), the surrogate model (middle), and combined base and surrogate (right). The experiment number index (exp) and cluster number (k) are annotated for a few selected significant experiments.*

As visualised in Figure , only a few clusters achieved a high score across mortality, stroke, and bleeding, however, we can easily find several experiments with at least two high scoring outcomes. Experiment 20 from ChelWest (k=6) cluster 5 is an example of a significant bleeding event, where the surrogate and combined scores remain high as well. We can also see how the surrogate model can affect the pattern screening; experiment 26 from ChelWest (k=5) cluster 3 on the other hand is an example that shows a dramatic decrease in bleeding score from the base model to the surrogate model. Though interestingly, the mortality score increases from base to surrogate. Note also, that by combining the original with the surrogate score, we still capture this experiment in the pattern screen. It is also noteworthy that the meta clustering experiments, such as C and E perform relatively poorly across the board.

*Surrogate Models of Heart Failure Experiments*

In some experiments surrogate models were able to produce simple, clinically interpretable decision trees which provided clear definitions for each cluster. This can be seen in the example from meta clustering experiment G (HF, ChelWest) shown in Figure 8. Whilst surrogate models replace the need to attempt to determine cluster definitions from complex, extensive enrichment tables, the simplified definitions provided by surrogate models can be understood in more depth by subsequent and focussed review of the enrichment table. For example, the definition of cluster 1 shown in Figure 8 could be further understood by reviewing the enrichment table which showed a higher prevalence in this cluster of co-morbidities such as respiratory and renal diseases which are associated with the procedure codes used in the surrogate models (Table 4).

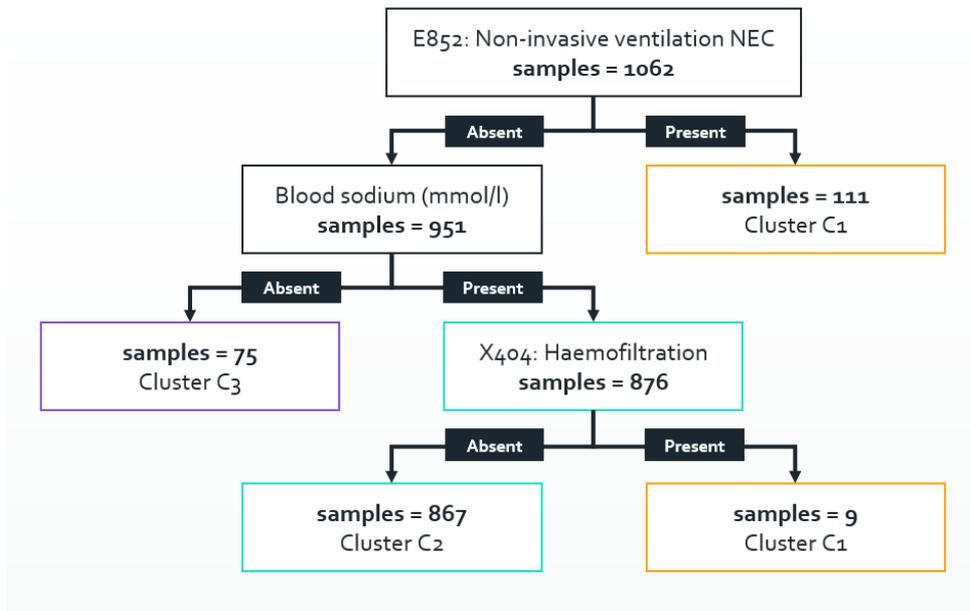

*Figure 8: Decision tree from Experiment G (HF, ChelWest).*

*Table 4: Truncated enrichment table showing some relevant features, such as E852 non-invasive ventilation NEC, which is enriched in cluster C1; blood sodium tests (p/a), which is under-enriched in cluster C3; X404 haemofiltration procedures, which is under-enriched in cluster C2, but enriched in cluster C1. Note that purple coloured text implies a positive odds ratio, and orange implies a negative odds ratio.*

| Feature | All Patients (freq. %) | Cluster 1 – 120 patients (freq. %) | Cluster 2 – 874 patients (freq. %) | Cluster 3 – 68 patients (freq. %) |
|---|---|---|---|---|
| **Diagnoses** | | | | |
| E872: Acidosis | 53 (5%) | 28 (23%) | 22 (2.5%) | 3 (4.4%) |
| J440: Chronic obstructive pulmonary disease with acute lower respiratory infection | 53 (5%) | 14 (12%) | 36 (4.1%) | 3 (4.4%) |
| J9600: Acute respiratory failure; Type I [hypoxic] | 20 (1.9%) | 9 (7.5%) | 8 (0.9%) | 3 (4.4%) |
| J969: Respiratory failure, unspecified | 43 (4%) | 30 (25%) | 13 (1.5%) | 0 |
| J9690: Respiratory failure, unspecified; Type I [hypoxic] | 37 (3.5%) | 14 (12%) | 23 (2.6%) | 0 |
| J9691: Respiratory failure unspecified; Type II [hypercapnic] | 39 (3.7%) | 29 (24%) | 10 (1.1%) | 0 |
| N179: Acute renal failure, unspecified | 277 (26%) | 50 (42%) | 214 (24%) | 13 (19%) |
| N185: Chronic kidney disease, stage 5 | 18 (1.7%) | 8 (6.7%) | 8 (0.9%) | 2 (2.9%) |
| | … | | | |
| **Procedures** | | | | |
| E852: Non-invasive ventilation NEC | 111 (10%) | 111 (92%) | 0 | 0 |
| X404: Haemofiltration | 20 (1.9%) | 20 (17%) | 0 | 0 |
| | … | | | |

The surrogate model's predictive performance was variable in practice, however in some cases the surrogate model was able predict the original clusters with a high degree of accuracy, even at a very low tree depth. For example, in meta clustering experiment G (HF, ChelWest) discussed above, the balanced accuracy of the surrogate model was 0.995 demonstrating successful prediction of the original clusters. Furthermore, both the original and surrogate model clusters showed significant differences for mortality between the clusters and therefore the clinical utility of the clusters had been preserved following application of the surrogate model.

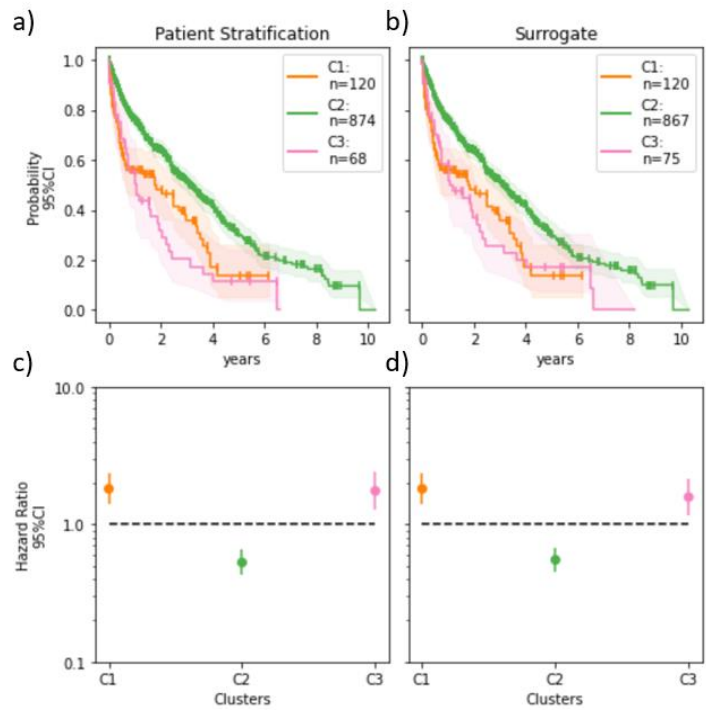

*Figure 9: Experiment G, ChelWest Heart Failure (HF) survival Kaplan Meier plot a) and b) with Cox PH model hazard ratios (HR) c) and d) for original patient stratification clusters (left) and surrogate model clusters (right). Note that Cox PH model is controlled for patient age and sex, and the log-rank p-values is less than 0.001 for both cases.*

*Cohort, Feature and Cluster Curation*

**Cohort curation** was used in experiment H (HF, OUH) where the clustering had identified a cluster of patients who lacked any blood tests during the period considered (Figure 10 left). This cluster was deemed irrelevant by clinical review as it was unlikely these patients were in acute heart failure; typically, blood tests are required as part of routine care. Further investigation revealed these patients frequently appeared to be admitted for day case procedures such as cardiac MRI, providing additional evidence against acute heart failure. This clustering analysis had therefore identified a previously unknown criteria which could be used to exclude patients from the cohort on the basis that they were highly unlikely to be in acute heart failure. A cohort curation was subsequently performed to remove these patients, thereby removing their influence on clustering, and the experiment was re-run (Figure 10 right). This process can be used to remove the undesirable influence of certain groups of patients identified through clustering as well as potentially adding new groups of patients. Importantly the removal of patients in this case was deemed clinically justifiable. Removal of patients could also occur based on evidence of bias producing irrelevant clusters.

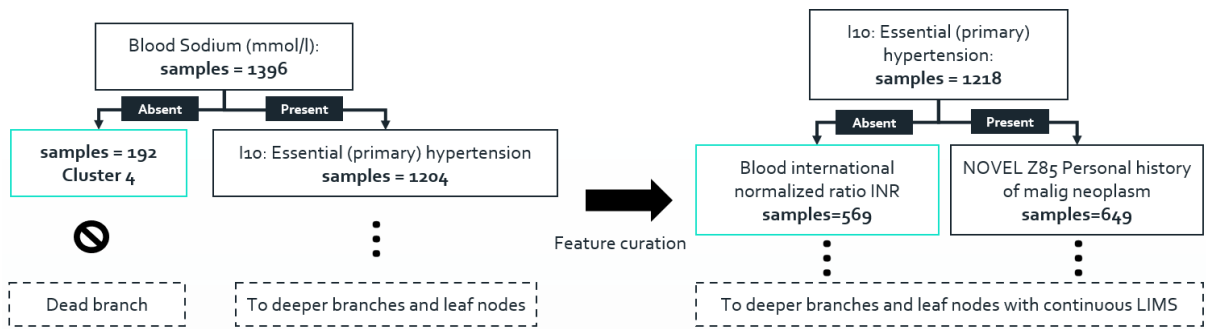

*Figure 10: Surrogate model decision tree from original Experiment H (left) and curated Experiment I (right) (HF, OUH) and after cohort and feature curation*

Figure 11 shows a sample of surrogate model without **feature curation** for experiment J (Stroke, OUH). Some of the features used by the model to define clusters, and appearing in the enrichment tables, in this case were discovered to be clinically irrelevant and could be removed completely, such as "Y981: Radiology of one body area (or <20 mins)". Further, several novel features with improved clinical relevance were created, for example: "Novel: Computed tomography angiography of cerebral vessels" was defined as "U212: Computed tomography NEC" AND ("Z342: Aortic arch" OR "Z35: Cerebral artery" OR "Z361: Carotid artery NEC"). Figure 12 shows a sample of the surrogate model for the same experiment post feature curation (Experiment K). All curated features are labeled as "Novel", which are defined in Appendix B. As visible in Figure 12, majority of features used within the surrogate model are curated features, demonstrating the ability to improve clinical relevance of features used to define clusters.

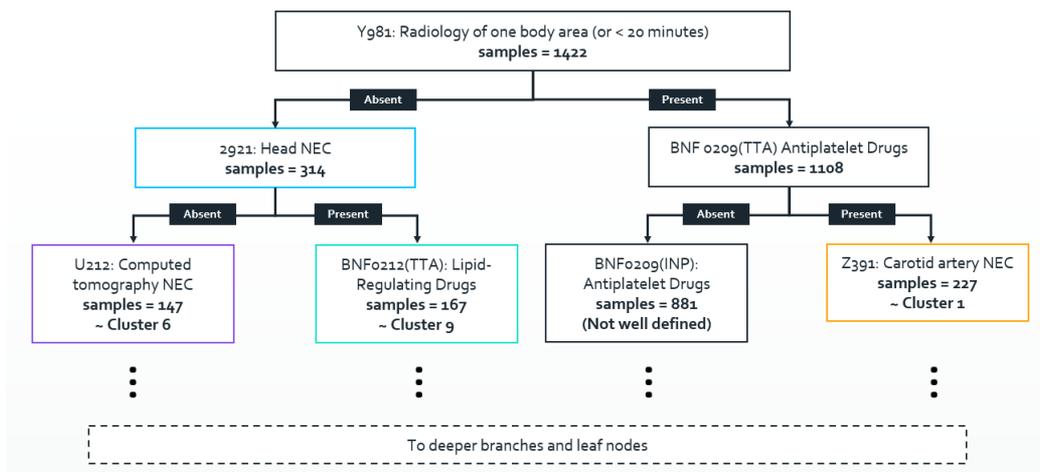

*Figure 11: First 3 levels of a surrogate model decision tree from experiment J (Stroke, OUH - uncurated).*

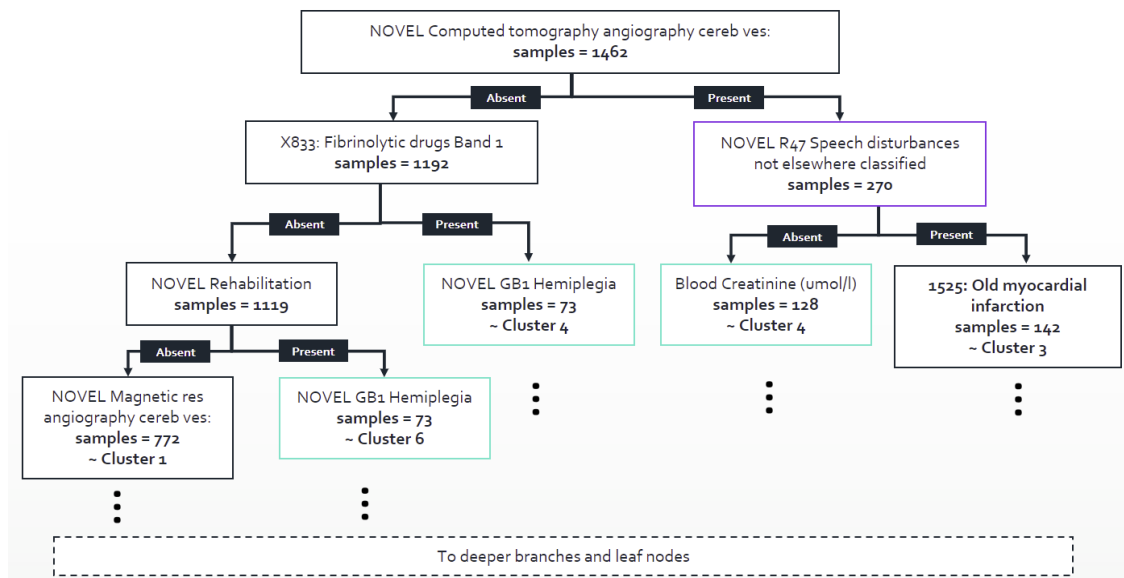

*Figure 12: Decision from experiment K (Stroke, OUH - curated).*

**Cluster curation** can also be demonstrated in this experiment. In the original model for this experiment, there were 7 clusters generated where cluster 7 exhibited an increase in mortality, see Figure 13a. However, the surrogate model failed to adequately capture and define this cluster (balanced accuracy 0.509). A cluster curation was therefore undertaken which compared patients within cluster 7 to the remaining cohort in attempt to simplify the surrogate model and obtain a definition for cluster 7, see Figure 13b. This increased the ability of the surrogate model to discriminate cluster 7 (balanced accuracy 0.659) from the remaining patients within the cohort. Unfortunately, the surrogate model predicted cluster 7 no longer showed a significant difference in mortality compared to the rest of the population. However, the cluster curation resulted in a simplified enrichment table when compared to the original experiment which could be used to define some features of cluster 7. These included enrichment for relevant co-morbidities such as atrial fibrillation, markers of stroke severity including gait/mobility issues and under enrichment for thrombolysis or thrombectomy, see Table 5.

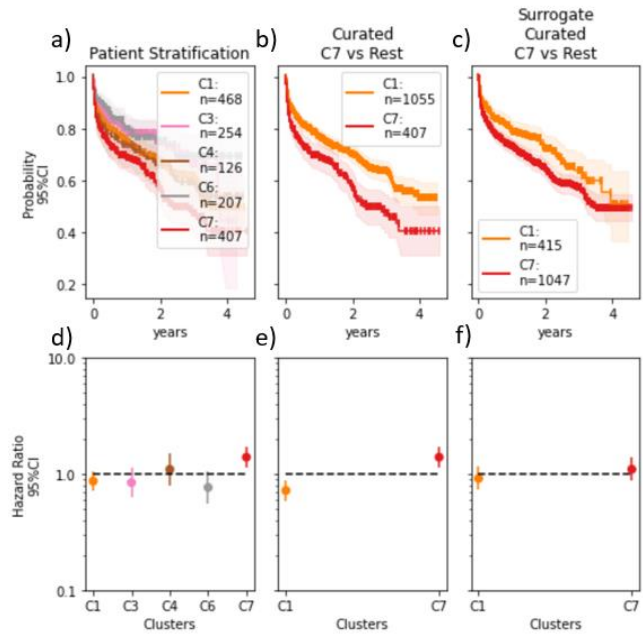

*Figure 13: Experiment J and K (OUH Stroke) survival Kaplan Meier (KM) plot for a) the patient stratification clusters from the original model clustering. The KM plot b) demonstrates the curated cluster C7 compared to the rest of the cohort, along with c) the complementary surrogate model. The Cox PH model hazard ratios for d) the original patient stratification clusters illustrate C7 as a cluster of interest with a significant hazard ratio after adjusting for age and sex. The e) curated C7 vs rest (new C1) shows the HRs are significant, however the f) curated surrogate model is unable to capture the same behaviour.*

*Table 5: Experiment K – C7 vs Rest (OUH stroke) selected enrichment table entries highlighting novel features.*

| Feature | All Patients (freq. %) | Cluster 1 – 1055 patients (freq. %) | Cluster 7 – 407 patients (freq. %) |
|---|---|---|---|
| **Diagnoses** | | | |
| NOVEL F05 delirium not induc alcohol or psychoact | 73 (4.8%) | 40 (3.8%) | 30 (7.4%) |
| NOVEL I48 Atrial fibrillation flutter | 522 (35%) | 327 (31%) | 187 (46%) |
| NOVEL I67 Other cerebrovascular diseases | 183 (12%) | 109 (10%) | 73 (18%) |
| NOVEL J18 Pneumonia organism unspecified | 100 (6.8%) | 55 (5.2%) | 42 (10%) |
| NOVEL M15 19 Arthrosis | 115 (7.8%) | 67 (6.4%) | 46 (11%) |
| NOVEL M81 Osteoporosis without path fract | 61 (4.1%) | 28 (2.7%) | 32 (7.9%) |
| NOVEL R26 Abnormalities of gait and mobility | 159 (11%) | 84 (8%) | 75 (18%) |
| R296: Tendency to fall, not elsewhere classified | 184 (12%) | 112 (11%) | 71 (17%) |
| | … | | |
| **Procedures** | | | |
| NOVEL Computed tomography angiography cereb ves | 277 (19%) | 269 (25%) | 1 (0.2%) |
| NOVEL Magn res angiography cereb ves | 27 (1.8%) | 25 (2.4%) | 0 |
| NOVEL Magnetic resonance imaging head | 115 (7.8%) | 90 (8.5%) | 21 (5.2%) |
| | … | | |

## Discussion

The increasing availability of EHRs has the potential to change many aspects of healthcare and drug development. Indeed, machine learning analysis of EHR data has great potential to produce insights into real-world patient journeys and identify novel patient subgroups. However, we have identified three key barriers to practical application, namely: the *identification* of clinically relevant results in a sea of analytical outputs, the *interpretation* of complex black box models, and model parameter and feature *optimization*. We have demonstrated, based on two (heart failure and stroke) patient stratification analyses use cases with real-world UK-based EHR data, tools for handling challenges in identifying patient clusters using meta clustering and pattern screening. Furthermore, we show the power of surrogate modelling for explaining patient phenotyping, and illustrate how feature, cluster, and cohort curation can be applied to optimize model results. The output of the use cases demonstrates how to produce condensed, prioritised, interpretable, and clinically relevant results. Significantly, this was achieved without compromising the technical advantages of deep learning based EHR analysis: complexity of models, parameters, and input options etc.

As we have shown, meta clustering allows for the rapid and simultaneous use of a wide range of models of varying complexity. There is a wealth of clustering algorithms at the disposal of a technically savvy researcher, and we do not suggest the specific algorithms we applied above are fit for all situations. It is often best practice to start with simpler models and build complexity to suit the application. For example, though not used in this study, a common baseline clustering method is k-Means clustering.(36) Additionally, this method can be supplemented by dimensionality reduction by converting the data into a compressed embedding via classical principal component analysis (PCA) for fitting k-Means. To build complexity, the models should suit the nature of the data; for example, sequential or time series data is well suited to modern transformers(37) or RNNs/LSTMs(38); image data have been classically handled using CNNs(39). In our case we have limited the number of models for demonstrative purposes, but we suggest that for further applications a wider selection of models can be used (not limited to deep learning).

There are several limitations to meta clustering. As shown in the meta clustering example applied to the heart failure cohort, we were able to reduce the number of initial reports by ~80%. However, it was not possible in all instances to find well defined meta clusters, such as with the stroke cohort using OUH data (see Appendix C). In this example, meta clustering was successfully applied to only a subset of results, namely from the DEC model. Additionally, meta clustering evaluation requires that all individual results use the same patient cohorts. Performing meta clustering analysis across multiple datasets is not possible. A further limitation of this approach is that only similar cluster assignments are considered and *not* the clinical outcomes. Both limitations can be overcome by using pattern screening.

Pattern screening is a simple yet effective way to handle large sets of analytical outputs and model reports. Not only can this approach handle multiple results across different models, but it can also compare different datasets and multiple outcomes. In our use case, we have shown it is effective in rapidly pinpointing significant results, e.g., targeted patient clusters.

Pattern screening is also highly flexible, where any number of metrics or scoring functions can be applied. Note that the pattern screening metrics we chose for the illustration are simplistic, and going forward, other metrics which not only consider the clinical questions but also the within cluster similarity might prove more discerning. The flexibility of pattern screening is underscored by the fact it can be used independently or in conjunction with meta clustering. Furthermore, in contrast to meta clustering where multiple results are combined and single unique results might be lost, pattern screening focuses on each individual cluster result. This is an important feature as there can be cases where a perceived outlier may in truth be a novel insight. This can also be seen in Figure 7, where though both individual as well as meta clustering results were ranked, clusters of meta clustering results seem to have lower pattern screening scores and therefore have less relevant clinical outcomes. This is sensible, as we are finding the consensus among a set of analyses, which independently of the clinical outcome. Additionally, meta clustering performed poorly in case of experiment C (see appendix C).

With this flexibility comes challenges. Indeed, pattern screening outputs depend strongly on how the criteria is defined, which means defining the criteria rationally is paramount. In our example the average scores for recurrent stroke appear to be deflated compared to mortality outcome. This is due the fact that the initial score is partly due to the low scores for the initial results compared to the lower surrogate model results (see exp 26 in Figure ). This indicates that the initial cluster did have a significantly outcome with respect to recurrent stroke, which was lost in the more generalized surrogate model. Again, this might be an informative outcome depending on the specific research questions.

We have used a simple decision tree for our surrogate model, however there are several well-developed options to choose from, such as Ripper(40), Trepan(41), or RuleFit(42) – the details of which are beyond the scope of this study. We have shown that even this simple decision tree surrogate model can be very useful in explaining deep learning clustering models. Certainly, the fact that the tree-based methods provide a list of important features automatically make it a very attractive option. This is particularly useful in the context of patient recruitment for clinical trials, as it can be used to generate relevant inclusion and exclusion criteria. However, there are several drawbacks to using tree-based methods, chiefly among them are the inability to handle temporal data. In our example, all temporal feature data was aggregated before it was applied to the surrogate model. In future work, there is scope to develop surrogate models that can accommodate patient trajectories.

It is worth noting that there is an inherent 'cost of explainability' when using surrogate models;(26) a black box model may achieve a higher level of accuracy than a white box or surrogate white box model, but in the field of medicine and clinical research, explainability is valued highly, and it is practical to trade accuracy for explainability. This is not necessarily a problem, but it must be understood and considered throughout the model development and interpretation process.

Of the optimisation steps discussed in this study, feature, cluster, and cohort curation undoubtably requires the most input from a domain expert. The creation of some novel features and the removal of others features can be intuitive in some cases for a clinical researcher and in other cases it is important to consider why irrelevant features may have been selected and what signal they represent from the data. An iterative approach to feature curation was used with different thresholds for combining and removing features to find a balance between searching for unbiased data driven insights versus ensuring clinical relevance of results. The choice of features to use is not trivial. On the one hand, it is critical to avoid confounding variables, which can cause an association to appear that doesn't exist. Many times this happens when an important variable is not controlled for – sometimes called a forking confounder – which distorts the association.(43) This can be achieved by adding all the features and "regressing out confounding effects from each input variable" before model training, or they can be controlled *post hoc*.(44) Yet there is a danger of blindly using every covariate available, which comes in the form of colliding confounders.(45)(46) A famous example of a colliding confounder is the obesity paradox in heart failure; in short, it has been found that patients stratified by mild to moderate obesity were associated with a decreased mortality risk.(47) Unsurprisingly, there is little evidence to suggest obesity has any *real* protective properties with respect to cardiovascular outcomes, and in fact has a negative impact on patient survival. To handle these types of biases requires domain knowledge, and these examples highlight the need for careful feature selection.

Curation is a somewhat subjective process. Based on training and experience, clinical researchers can identify meaning – or lack thereof – in EHR signals. This, for example, applies to the relevance of features for individual patients (e.g., discarding a laboratory assessment as not relevant for a specific patient), but also for definitions of a patient cohort. Such definitions should ideally be simple to apply and concentrate on medically meaningful concepts that relate to the condition or outcome of interest. These concepts are difficult to implement in an unbiased way and usually requires expert input. Our approach to curation at different levels of analysis (cohort, feature, and cluster) therefore aimed to make it easy to obtain this input from clinical experts by reducing definition criteria and make their meaning explicit.

Cohort curation has some overlap with data quality analysis and cleaning. Plainly, any spurious patient data should be omitted, e.g., male patients in a gestational diabetes cohort or female patients in a prostate cancer cohort. However, cohort curation should also ensure that the patients represent the target population. This implies that there should ideally be a balance between sexes (if not a sex dependent disease), as well as a distribution of patient ages that correspond to the target population. It is also important that underrepresented ethnicities are not lost. And in some cases, these populations should be enriched, especially if they represent an outsized burden for a disease area. In the context of clinical trials, this has been clearly outlined in a 2022 USA FDA draft guidance for industry.(48)

The primary objective of this study was to identify subpopulations within a larger sample based on patient characteristics in their EHRs. Indeed, the utility of defining patient clusters with clinically relevant features is underscored by the need to find suitable data driven clinical trial criteria. A natural potential application of the methods outlined in this study would be to use an understanding of patient subgroups to guide selection for clinical trials, allowing for patients of different profiles to be recruited based on prior diagnoses, blood test results or medical procedures. Moreover, the subgroups produced by the clustering algorithm can support targeted recruitment of patients for clinical trials who are likely to be enriched with the desired outcome such as mortality, recurrence of disease or secondary conditions.

In addition to clinical trial optimisation, this scalable clinical interpretation approach can be applied to fundamental research of a disease area. By pin-pointing sub-phenotypes, we can better understand disease progression for a richer variety of patients. And in the future, with more widespread genomic and transcriptomic testing matched to full patient EHRs, we can drastically increase our understanding of disease and improve therapy selection.(49)(50)

We have shown in depth the strength of the approach through a use case and have highlighted the key techniques developed. However, we have found that there are several limitations inherent to the process, which include limits in the data itself, the use and misuse of the process. Despite recent advances, it is still common to find EHR datasets which are incomplete, with missing documentation for patients from general practitioners, imaging diagnostic, or surgeries. Furthermore, limitations may arise impacting longitudinal aspects, e.g., data only being available for a certain period making it unfeasible to observe both health, start, and advanced stages for many chronic diseases in the same patient. Moreover, the breadth and quality of the data will naturally affect the feasibility of the proposed scalable clinical interpretation models.

Additionally, though more of a caveat than a limitation, we do not propose a replacement for human analysis, rather a tool for optimised human-in-the-loop analysis. A corollary of this is the necessity of a domain expert to define the models and metrics. For example, our current approach uses a silhouette score that automatically determines the optimal number of meta cluster $k$. However, this formulation may not be well suited to answer any given clinical questions as it only considers the average width of a cluster and its distance to the nearest neighbouring cluster. Metrics such as these still require both clinical and technical expertise for this approach to be successful.

In summary, we have developed methods that are emulating parts of an 'intuitive' approach a clinical researcher may choose to take when reviewing EHR analyses. These methods not only increase clinical meaning and facilitate throughput of the analyses, but they also provide a common language for data scientists and clinicians to support collaboration in an interactive approach. With this we expect the methods presented here, and future improvement based on this, to substantially increase learnings and help unlock the potential of real-world data in improving clinical practice and focusing drug development efforts.

# Acknowledgements

We thank Oxford University Hospitals and Chelsea and Westminster Hospitals trust for access to anonymized data, as well as Sam Allen for designing our figures and charts. Additionally, we thank specifically Christian Diedrich, Eren Elci, Steffen Schaper, Basel Abu-Jamous, Niklas Kokkola, Fernando Andreotti and all Sensyne and Bayer members of this research project.

This work uses data provided by patients collected by Chelsea and Westminster Hospital NHS Foundation Trust and Oxford University Hospitals NHS Foundation Trust as part of their care and support. We believe using the patient data is vital to improve health and care for everyone and would, thus, like to thank all those involved for their contribution. The data were extracted, anonymized, and supplied by the Trust in accordance with internal information governance review, NHS Trust information governance approval, and the General Data Protection Regulation (GDPR) procedures outlined under the Strategic Research Agreement (SRA) and relative Data Processing Agreements (DPAs) signed by the Trust and Sensyne Health plc.

This research has been conducted using the Oxford University Hospitals NHS Foundation Trust Clinical Data Warehouse, which is supported by the NIHR Oxford Biomedical Research Centre and Oxford University Hospitals NHS Foundation Trust. Special thanks to Kerrie Woods, Kinga Varnai, Oliver Freeman, Hizni Salih, Steve Harris and Professor Jim Davies.

# Footnote

*Conflicts of Interest*: The authors have no conflicts of interest to declare.

*Ethical Statement*: The authors are accountable for all aspects of the work in ensuring that questions related to the accuracy or integrity of any part of the work are appropriately investigated and resolved.

*Open Access Statement:* This is an Open Access article distributed in accordance with the Creative Commons Attribution-NonCommercial-NoDerivs 4.0 International License (CC BY-NC-ND 4.0), which permits the non-commercial replication and distribution of the article with the strict proviso that no changes or edits are made and the original work is properly cited (including links to both the formal publication through the relevant DOI and the license).

# References

1. Leonelli S. Data-centric biology. University of Chicago Press; 2016.
2. Wiens J, Shenoy ES. Machine Learning for Healthcare: On the Verge of a Major Shift in Healthcare Epidemiology. Clin Infect Dis. 2018;66(1):149–53.
3. Beckmann JS, Lew D. Reconciling evidence-based medicine and precision medicine in the era of big data: Challenges and opportunities. Genome Med. 2016;8(1).
4. Collins FS, Varmus H. A New Initiative on Precision Medicine. N Engl J Med. 2015;372(9):793–5.
5. Parimbelli E, Marini S, Sacchi L, Bellazzi R. Patient similarity for precision medicine: A systematic review. J Biomed Inform. 2018;83:87–96.
6. Javer A, Parsons O, Carr O, Baxter J, Diedrich C, Elçi E, et al. Compensating trajectory bias for unsupervised patient stratification using adversarial recurrent neural networks. arXiv Prepr arXiv211207239. 2021;
7. Chen R, Sun J, Dittus RS, Fabbri D, Kirby J, Laffer CL, et al. Patient Stratification Using Electronic Health Records from a Chronic Disease Management Program. IEEE J Biomed Heal Informatics. 2020;1.
8. Hedman ÅK, Hage C, Sharma A, Brosnan MJ, Buckbinder L, Gan LM, et al. Identification of novel pheno-groups in heart failure with preserved ejection fraction using machine learning. Heart. 2019;
9. Kobayashi M, Huttin O, Magnusson M, Ferreira JP, Bozec E, Huby A-C, et al. Machine Learning-Derived Echocardiographic Phenotypes Predict Heart Failure Incidence in Asymptomatic Individuals. JACC Cardiovasc Imaging. 2022;15(2):193–208.
10. Parker JS, Mullins M, Cheang MCU, Leung S, Voduc D, Vickery T, et al. Supervised Risk Predictor of Breast Cancer Based on Intrinsic Subtypes. J Clin Oncol. 2009;27(8):1160–7.
11. Segar MW, Patel K V, Ayers C, Basit M, Tang WHW, Willett D, et al. Phenomapping of patients with heart failure with preserved ejection fraction using machine learning-based unsupervised cluster analysis. Eur J Heart Fail. 2020;22(1):148–58.
12. Uijl A, Savarese G, Vaartjes I, Dahlström U, Brugts JJ, Linssen GCM, et al. Identification of distinct phenotypic clusters in heart failure with preserved ejection fraction. Eur J Heart Fail. 2021;23(6):973–82.
13. Guillamet RV, Ursu O, Iwamoto G, Moseley PL, Oprea T. Chronic obstructive pulmonary disease phenotypes using cluster analysis of electronic medical records. Health Informatics J. 2018;24(4):394–409.
14. Vranas KC, Jopling JK, Sweeney TE, Ramsey MC, Milstein AS, Slatore CG, et al. Identifying Distinct Subgroups of ICU Patients. Crit Care Med. 2017;45(10):1607–15.
15. Williams JB, Ghosh D, Wetzel RC. Applying Machine Learning to Pediatric Critical Care Data*. Pediatr Crit Care Med. 2018;19(7):599–608.
16. Karwath A, Bunting K V, Gill SK, Tica O, Pendleton S, Aziz F, et al. Redefining β-blocker response in heart failure patients with sinus rhythm and atrial fibrillation: a machine learning cluster analysis. Lancet. 2021;398(10309):1427–35.
17. Wang Y, Zhao Y, Therneau TM, Atkinson EJ, Tafti AP, Zhang N, et al. Unsupervised machine learning for the discovery of latent disease clusters and patient subgroups using electronic health records. J Biomed Inform. 2020;102:103364.
18. MacQueen J, others. Some methods for classification and analysis of multivariate observations. In: Proceedings of the fifth Berkeley symposium on mathematical statistics and probability. 1967. p. 281–97.
19. Carr O, Javer A, Rockenschaub P, Parsons O, Durichen R. Longitudinal patient stratification of electronic health records with flexible adjustment for clinical outcomes. In: Roy S, Pfohl S, Rocheteau E, Tadesse GA, Oala L, Falck F, et al., editors. Proceedings of Machine Learning for Health [Internet]. PMLR; 2021. p. 220–38. (Proceedings of Machine Learning Research; vol. 158). Available from: https://proceedings.mlr.press/v158/carr21a.html
20. Hothorn T, Hornik K, Zeileis A. ctree: Conditional inference trees. Compr R Arch Netw. 2015;8.
21. Monti S, Tamayo P, Mesirov J, Golub T. Consensus clustering: a resampling-based method for class discovery and visualization of gene expression microarray data. Mach Learn. 2003;52(1):91–118.
22. Aure MR, Vitelli V, Jernström S, Kumar S, Krohn M, Due EU, et al. Integrative clustering reveals a novel split in the luminal A subtype of breast cancer with impact on outcome. Breast Cancer Res. 2017;19(1):1–18.
23. Müllner D. Modern hierarchical, agglomerative clustering algorithms. arXiv Prepr arXiv11092378. 2011;
24. Rousseeuw PJ. Silhouettes: a graphical aid to the interpretation and validation of cluster analysis. J Comput Appl Math. 1987;20:53–65.
25. Loyola-Gonzalez O. Black-box vs. white-box: Understanding their advantages and weaknesses from a practical point of view. IEEE Access. 2019;7:154096–113.
26. Goethals S, Martens D, Evgeniou T. The non-linear nature of the cost of comprehensibility. J Big Data. 2022;9(1):1–23.
27. Caruana R, Elhaway M, Nguyen N, Smith C. Meta clustering. In: Sixth International Conference on Data Mining (ICDM'06). 2006. p. 107–18.
28. Srivastava M. A Surrogate data-based approach for validating deep learning model used in healthcare. In: Applications of Deep Learning and Big IoT on Personalized Healthcare Services. IGI Global; 2020. p. 132–46.
29. Barmada S, Fontana N, Formisano A, Thomopulos D, Tucci M. A deep learning surrogate model for topology optimization. IEEE Trans Magn. 2021;57(6):1–4.

30. Thakur A, Chakraborty S. A deep learning based surrogate model for stochastic simulators. Probabilistic Eng Mech. 2022;68:103248.
31. Pennington J, Socher R, Manning CD. Glove: Global vectors for word representation. In: Proceedings of the 2014 conference on empirical methods in natural language processing (EMNLP). 2014. p. 1532–43.
32. Hong C, Kim H, Oh J, Lee KM. DAQ: Distribution-Aware Quantization for Deep Image Super-Resolution Networks. arXiv Prepr arXiv201211230. 2020;
33. Xie J, Girshick R, Farhadi A. Unsupervised deep embedding for clustering analysis. In: International conference on machine learning. 2016. p. 478–87.
34. Freudenberg JM, Joshi VK, Hu Z, Medvedovic M. CLEAN: CLustering Enrichment ANalysis. BMC Bioinformatics. 2009;10(1):234.
35. Davidson-Pilon C. lifelines: survival analysis in Python. J Open Source Softw [Internet]. 2019;4(40):1317. Available from: https://doi.org/10.21105/joss.01317
36. Horiuchi Y, Tanimoto S, Latif AHMM, Urayama KY, Aoki J, Yahagi K, et al. Identifying novel phenotypes of acute heart failure using cluster analysis of clinical variables. Int J Cardiol. 2018;262:57–63.
37. Vaswani A, Shazeer N, Parmar N, Uszkoreit J, Jones L, Gomez AN, et al. Attention is all you need. Adv Neural Inf Process Syst. 2017;30.
38. Graves A. Long short-term memory. Supervised Seq Label with Recurr neural networks. 2012;37–45.
39. Sermanet P, Chintala S, LeCun Y. Convolutional neural networks applied to house numbers digit classification. In: Proceedings of the 21st international conference on pattern recognition (ICPR2012). 2012. p. 3288–91.
40. Cohen WW. Fast effective rule induction. In: Machine learning proceedings 1995. Elsevier; 1995. p. 115–23.
41. Craven M, Shavlik J. Extracting tree-structured representations of trained networks. Adv Neural Inf Process Syst. 1995;8.
42. Friedman JH, Popescu BE. Predictive learning via rule ensembles. Ann Appl Stat. 2008;2(3):916–54.
43. JK A, C B, D N. Catalogue of Bias Collaboration: Confounding. Cat Bias [Internet]. 2018; Available from: www.catalogueofbiases.org/biases/confounding
44. Dinga R, Schmaal L, Penninx BWJH, Veltman DJ, Marquand AF. Controlling for effects of confounding variables on machine learning predictions. BioRxiv. 2020;
45. H L, JK AJ, D N. Catalogue of Bias Collaboration: Collider. Cat Bias [Internet]. 2019; Available from: https://catalogofbias.org/biases/collider-bias/
46. Sackett DL. Bias in analytic research. In: The case-control study consensus and controversy. Elsevier; 1979. p. 51–63.
47. Clark AL, Fonarow GC, Horwich TB. Obesity and the obesity paradox in heart failure. Prog Cardiovasc Dis. 2014;56(4):409–14.
48. Diversity Plans to Improve Enrollment of Participants From Underrepresented Racial and Ethnic Populations in Clinical Trials. 2022; Available from: https://www.fda.gov/media/157635/download
49. Docking TR, Parker JDK, Jädersten M, Duns G, Chang L, Jiang J, et al. A clinical transcriptome approach to patient stratification and therapy selection in acute myeloid leukemia. Nat Commun. 2021;12(1):1–15.
50. Fujio K, Takeshima Y, Nakano M, Iwasaki Y. transcriptome and trans-omics analysis of systemic lupus erythematosus. Inflamm Regen. 2020;40(1):1–6.

# Appendix

*Appendix A: Outcome definition*

In a similar manner to the definition of mortality, the specific definitions of the outcomes: recurrence of stroke, bleeding events, and readmission for heart failure are provided.

Recurrence of stroke

Presence of repeat hospitalisation for ischaemic stroke (as defined below) following index hospitalisation for ischaemic stroke. Admission to an inpatient unit or emergency department with any of the following primary ICD10 codes:
- I63* Cerebral infarction

EXCLUDING:
Any admissions with the admission code:
- 81- Transfer of any admitted PATIENT from other Hospital Provider other than in an emergency (NHS Data Dictionary)

Bleeding events

Any of the following primary ICD10 codes:
- K25.0 Acute gastric ulcer with haemorrhage
- K25.2 Acute gastric ulcer with both haemorrhage and perforation
- K25.4 Chronic or unspecified gastric ulcer with hemorrhage
- K25.6 Chronic or unspecified gastric ulcer with both hemorrhage and perforation
- K26.0 Acute duodenal ulcer with hemorrhage
- K26.2 Acute duodenal ulcer with both hemorrhage and perforation
- K26.4 Chronic or unspecified duodenal ulcer with hemorrhage
- K26.6 Chronic or unspecified duodenal ulcer with both hemorrhage and perforation
- K27.0 Acute peptic ulcer, site unspecified, with hemorrhage
- K27.2 Acute peptic ulcer, site unspecified, with both hemorrhage and perforation
- K27.4 Chronic or unspecified peptic ulcer, site unspecified, with hemorrhage
- K27.6 Chronic or unspecified peptic ulcer, site unspecified, with both hemorrhage and perforation
- K28.0 Acute gastrojejunal ulcer with hemorrhage
- K28.2 Acute gastrojejunal ulcer with both hemorrhage and perforation
- K28.4 Chronic or unspecified gastrojejunal ulcer with hemorrhage
- K28.6 Chronic or unspecified gastrojejunal ulcer with both hemorrhage and perforation
- K92.0 Hematemesis
- I60* Subarachnoid hemorrhage
- S06.6 Traumatic subarachnoid hemorrhage
- I62.0 Nontraumatic subdural hemorrhage
- S06.5 Traumatic subdural hemorrhage
- I61* Intracerebral hemorrhage
- I62.1 Nontraumatic extradural hemorrhage
- I62.9 Nontraumatic intracranial hemorrhage, unspecified
- K92.1 Melena
- I85.0 Oesophageal varices with bleeding
- I98.3 Oesophageal varices with bleeding in diseases classified elsewhere

OR

Administration of any of the following medications:
- Dried prothrombin complex
- Fresh frozen plasma
- Idarucizumab

OR

Administration of any of the following procedures (OPCS4):
- G20.1 Fibreoptic endoscopic coagulation of bleeding lesion of oesophagus
- G46.2 Fibreoptic endoscopic coagulation of bleeding lesion of upper gastrointestinal tract
- X33.2 Intravenous blood transfusion of packed cells

Readmission for heart failure

Inpatient admission or emergency department admissions 2+ days in length which meet the following criteria:
- A primary ICD10 code for HF defined as:
  - I50* Heart failure
  - I11.0 Hypertensive heart disease with heart failure
  - I13.0 Hypertensive heart and renal disease with (congestive) heart failure
  - I13.2 Hypertensive heart and renal disease with both (congestive) heart failure and renal failure

OR
- A secondary ICD10 code for HF as (defined above) AND IV Furosemide administered within admission or within 1 day following admission

EXCLUDING

- Any admission that is less than 48 hours during which one of the following procedures is performed:
  - Cardioverter defibrillator introduced through the vein (OPCS-4: K59*)
  - Other cardiac defibrillator (OPCS-4: K72*)
  - Cardiac pacemaker system introduced through vein (OPCS-4: K60*)
  - Other cardiac pacemaker system (OPCS-4: K61*)
  - Other cardiac pacemaker system introduced through vein (OPCS-4: K73*)
  - Cardiac pacemaker system (OPCS-4: K74*)

*Appendix B: Novel Feature definitions*

The specific novel feature definitions that are discussed in this paper are provided here. Each line is connected by an 'AND' or 'OR' statement, which can be further connected by an indentation of the same logic.

**Computed tomography angiography of cerebral vessels:**
U212: Computed tomography NEC
AND
(
   Z342: Aortic arch
   OR
   Z35: Cerebral artery OR Z361: Carotid artery NEC
)

**Speech disturbances not elsewhere specified:**
All R47 subcodes combined

**Rehabilitation:**
Z501: Other physical therapy
OR
Z505: Speech therapy
OR
Z507: Occupational therapy and vocational rehabilitation, not elsewhere classified

**Hemiplegia:**
All G81 subcodes combined

**Magnetic resonance angiography of cerebral vessels:**
U211: Magnetic resonance imaging NEC
AND
(
   Z342: Aortic arch
   OR
   Z35: Cerebral artery OR Z361: Carotid artery NEC
)

*Appendix C: Example of poor performance of meta clustering*

Experiment C, OUH Stroke with k=5 is an example of a meta clustering analysis that did not perform well. In contrast to the example in Figure 6 where we see good separation with distinct blocks of patients in the dendrogram, there is no clear consensus of which clustering the patients should belong. Note that the Kaplan Meier and Cox plots are provided purely for illustrative purposes, meta clustering does not optimise the separation of outcomes.

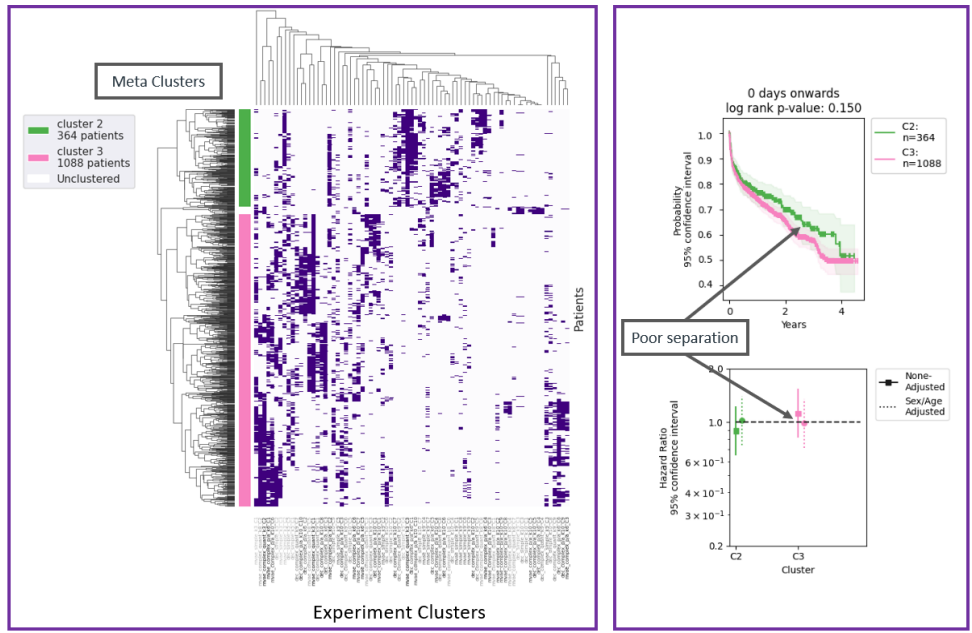

*Figure S1: Experiment C, an example of poor performance of meta clustering (left) with a dendrogram heatmap and (right) survival analysis. Here the patient groups do not show significantly different outcomes.*

Page 27 of 27